\documentclass{article}

\usepackage[preprint, nonatbib]{neurips_2026} 
\usepackage{cite}
\usepackage{graphicx} 
\usepackage{amsmath,amssymb,amsfonts}
\usepackage[utf8]{inputenc} 
\usepackage[T1]{fontenc}    
\usepackage{hyperref}       
\usepackage{url}            
\usepackage{booktabs}      
\usepackage{nicefrac}       
\usepackage{microtype}      
\usepackage{xcolor} 
\usepackage{multirow}

\title{
Improving the Robustness/Accuracy Tradeoff Against Adversarial Attacks Using Information Bottleneck Distillation Through Dual Teachers
}
\author{
    Ryusuke V.~Takahashi \\
    Joint Graduate School of Mathematics for Innovation\\
    Kyushu University\\
    \texttt{takahashi@me.inf.kyushu-u.ac.jp}
    \And
    Yoshinari~Takeishi \\
    Faculty of Information Science and Electrical Engineering\\
    Kyushu University\\
    \texttt{takeishi@inf.kyushu-u.ac.jp}
    \And
    Jun'ichi~Takeuchi \\
    Faculty of Information Science and Electrical Engineering \\
    Kyushu University \\
    \texttt{tak@inf.kyushu-u.ac.jp}
    \And
    Kavé Salamatian \\
    Université Savoie Mont Blanc \\
    \texttt{kave.salamatian@univ-smb.fr}
}

\begin{document}

\maketitle
\begin{abstract}

Deep neural networks (DNNs) have achieved remarkable success in classical machine learning problems. However, they are known to be vulnerable to adversarial attacks. Countermeasures proposed in the literature, notably Information Bottleneck Distillation (IBD) introduced by Kuang  {\em et al.}, degrade the classification accuracy on clean inputs while improving the robustness to adversarial inputs. In this work, we extend the IBD framework by introducing an extra teacher model (clean teacher) trained with only clean inputs, into the distillation process from a robust teacher model trained by adversarial training. The features of both clean and robust teachers are transferred to the student through a cross-layer attention matrix. Experimental results on the CIFAR-10 and CIFAR-100 datasets show that the proposed method improves classification accuracy on clean samples compared to the original IBD, while maintaining similar accuracy on adversarial samples. Furthermore, our methods are competitive with state-of-the-art approaches, including the recent dual-teacher distillation framework B-MTARD, particularly in terms of the harmonic mean between clean and robust accuracy. We also analyze the impact of different training settings that have different influences on the attention module. Our code is available at \url{https://anonymous.4open.science/r/DD_JD-9791}.
\end{abstract}

\section{Introduction}

\label{intro}

Deep neural networks (DNNs) have been shown to perform extremely well on many classical machine learning problems. However, deep learning is vulnerable to adversarial samples, {\em i.e.}, crafted input designed to fool the model into making an incorrect prediction \cite{Goodfellow2015}. Such attempts seriously undermine the safety and reliability of DNNs for real-world applications. Extensive research, such as adversarial training \cite{Madry2018} and its many variants, has as its main objective to enhance the adversarial robustness of DNN models. Following the proposition in \cite{Hinton2015}, a knowledge distillation using the Information Bottleneck (IB) principle, named Information Bottleneck Distillation (IBD), was proposed in \cite{Kuang2023}. In this approach, a student model is trained using the output of a larger, robust teacher model. In this paper, we extend this approach by proposing a distillation framework leveraging two distinct teacher models. One, the ``{\em robust teacher}", is similar to the one proposed in \cite{Kuang2023}; a second, the ``{\em clean teacher}", specializes in clean sample classification. This design is expected to preserve robustness against adversarial samples while improving classification performance on clean ones. We have studied two strategies for distilling intermediate features from the teacher models to the student: \textbf{Double distillation}, in which the student model aligns its intermediate representations with those of the robust teacher alone (Figure \ref{fig:twodist}), and \textbf{Joint distillation}, in which the student model distills knowledge from both the robust and the clean teachers simultaneously (Figure \ref{fig:jointdist}).

We further investigate the behavior of the attention module under two training setups. In one setup, the learned attention weights become nearly uniform, with values close to the reciprocal of the number of student features. In the other, they become highly concentrated, with dominant weights close to one. We show empirically that the proposed methods outperform the baseline provided by IBD in clean accuracy while preserving a similar robust accuracy. Furthermore, our methods are competitive with state-of-the-art methods, including B-MTARD \cite{Zhao2024}, a recent dual-teacher distillation framework for adversarial robustness, particularly in terms of the harmonic mean between clean and robust accuracy. We found that the proposed methods are particularly effective for larger models, and the magnitude of the improvement is comparable to the gains reported in recent papers such as \cite{Kuang2023} and \cite{Zhang2024}.

\section{Background \& Related Work}

\subsection{Adversarial attacks and robustness against them}

With deep neural networks becoming more popular, the integrity of the applications that leverage them becomes more crucial. The authors of  \cite{Szegedy2014} showed that attackers can manipulate valid inputs, imperceptibly for human eye, and force a trained DNN model to generate incorrect outputs. There is already a rich literature on the effects of adversarial attacks \cite{PAWLICKI2025131231}, and \cite{Shafahi2019} about attacks on autonomous vehicles. It is noteworthy that generated adversarial samples are also used as a way improving the robustness of Machine Learning approaches, leading to joint attack-defense training frameworks. In \cite{Pelekis2025}, a survey of current machine learning mitigation techniques is provided. This survey underline that adversarial training, {\em i.e.}, improving the training set by adding samples tempered by an adversarial attacker, that was introduced in \cite{Shafahi2019}, remains the strongest baseline defense approach toward adversarial attacks. While these approaches have been shown to decrease the impact of adversarial attacks, using them degrades the performance of the model. The aim of this paper is to investigate this trade-off in the context of the Information Bottleneck method.

\subsection{Knowledge Distillation}

Knowledge distillation, proposed in \cite{Hinton2015}, consists of a small ``\textit{student}" model that is trained to mimic the behavior of a larger and more complex ``\textit{teacher}" model, by learning the output probabilities (soft outputs) of teacher. Distillation have initially been shown to reduce sensitivity to gradient-based adversarial attacks \cite{Papernot2016}. Nonetheless, Carlini and Wagner designed stronger attacks, named CW attacks, that were not mitigated by distillation \cite{Carlini2016}. This has recently led to researches \cite{Goldblum2020, Zi2021} showing that while adversarial training is the real source of robustness to adversarial attacks, distillation plays the role of efficiently transferring the robustness to the student model.
In addition to aligning the student’s output with the teacher’s soft labels, some methods also encourage the student to mimic the internal representations of specific intermediate layers of the teacher model \cite{Romero2015}. Knowledge distillation has been shown to enhance robustness against adversarial examples as well \cite{Papernot2016}. More recently, B-MTARD \cite{Zhao2024} proposed a two-teacher distillation framework for adversarial robustness, showing that complementary teacher models can improve the balance between clean and robust accuracy.

\subsection{Information Bottleneck Method}

The Information Bottleneck (IB) method \cite{Tishby2000} was introduced as a principle for extracting the most relevant parts of an input variable $X$ for predicting an output variable $Y$, while compressing the input as much as possible. We want to find a compressed representation $Z=\mathcal{T}(X)$ such that $Z$ retains as much as possible information about $Y$, and $\mathcal{T}(X)$ uses as little as possible of the information of $X$.  This leads to a balance between compression and relevance that is formulated as the following optimization problem:

\begin{align}
\max_{p(z|x)}\ I(Z; Y) - \beta I(X; Z).
\label{eq:IB}
\end{align}

where optimization is over all conditional distributions $p(z|x)$. The parameter $\beta > 0$ controls the trade-off between relevance (maximizing $I(Z; Y)$) and compression (minimizing $I(X; Z)$).


This method can be used in training DNNs \cite{Tishby2015, Saxe2018}. However, a major drawback of this method is the difficulty of computing the two mutual information in Eq. \eqref{eq:IB}. Several approaches have been proposed to address this issue, which will be further discussed in Section \ref{subsec:IBD}.

\subsection{Information Bottleneck Distillation}
\label{subsec:IBD}

The loss function used in classical distillation, $\mathcal{L}_{\text{distill}}$ combines two terms:

\begin{equation}
\label{eq:distill}
\begin{split}
\mathcal{L}_{\text{distill}}
&= \gamma \cdot \mathcal{L}_{\text{CE}}
\bigl(y_{\text{student}}, y_{\text{true}}\bigr) \\
&\quad + (1 - \gamma) \cdot \mathcal{L}_{\text{KL}}
\bigl(y_{\text{student}}, y_{\text{teacher}}\bigr)
\end{split}
\end{equation}

Where $\mathcal{L}_{\text{CE}}$ is the cross-entropy loss with ground truth labels, $\mathcal{L}_{\text{KL}}$ is the  Kullback-Leibler (KL) divergence between student and teacher soft outputs, and $0<\gamma<1$ is a trade-off parameter balancing the two terms.
Following the suggestion made in \cite{Hinton2015, Alemi2016, Fischer2020}, the IBD \cite{Kuang2023} structure uses information from a teacher model to become robust against adversarial attacks. Rather than computing the mutual information, IBD relies on knowledge distillation to approximate the IB objective. Specifically, this method uses the cross-entropy between the teacher soft labels and the student output for the relevance term, and an attention-weighted difference of the latent features of the teacher and student models for the compression term. The method uses a matrix of weights ($\text{Attn}_{i,j}$), called ``attention matrix", to link the teacher feature of layer $i$ to the student feature in layer $j$. Combined with adversarial training, this results in the following loss function originally defined in \cite{Kuang2023}:

\begin{equation}
\begin{split}
\mathcal{L}_{\text{IBD-adv}} = \mathbb{E}_{p(x_{\text{nat}})} \Big[ & 
(1 - \alpha)\, \mathcal{L}_{\text{CE}}(h_s(x_{\text{nat}}), y_t) \\
& + \alpha \, \mathcal{L}_{\text{CE}}(h_s(x_{\text{adv}}), y_t) \\
& + \beta \sum_{i=1}^{n} \sum_{j=1}^{m} \text{Attn}_{i,j} 
\big(T_t^i(z_t^i(x_{\text{adv}})) \\
& - T_s^j(z_s^j(x_{\text{adv}}))\big)^2
\Big]
\end{split}
\label{eq:IBDadv}
\end{equation}

\noindent where $\alpha$ is a trade-off parameter between training on the non-adversarial (clean) inputs $x_\text{nat}$,  and  the  adversarial (robust) input $x_{\text{adv}}$. T denotes the feature transformation applied before feature matching to make teacher and student feature representations (respectively $z_t$ and $z_s$) directly comparable.

\section{Double and Joint IBD architecture}

A major issue for adversarial robustness is the tradeoff between clean and adversarial accuracy. The literature reports that increasing robustness entails a decrease in clean data accuracy. The aim of this paper is to improve this trade-off. Our intuition behind these two proposed architectures is that the use of the clean teacher should improve the performance of clean inputs in the distilled student. For this purpose, we are proposing two new architectures for Information Bottleneck Distillation. Both approaches involve using a ``clean teacher" working only on non-adversarial input, in addition to the adversarial ``robust teacher" originally used in \cite{Kuang2023}.

\begin{figure}
    \centering
    \includegraphics[width=0.8\columnwidth]{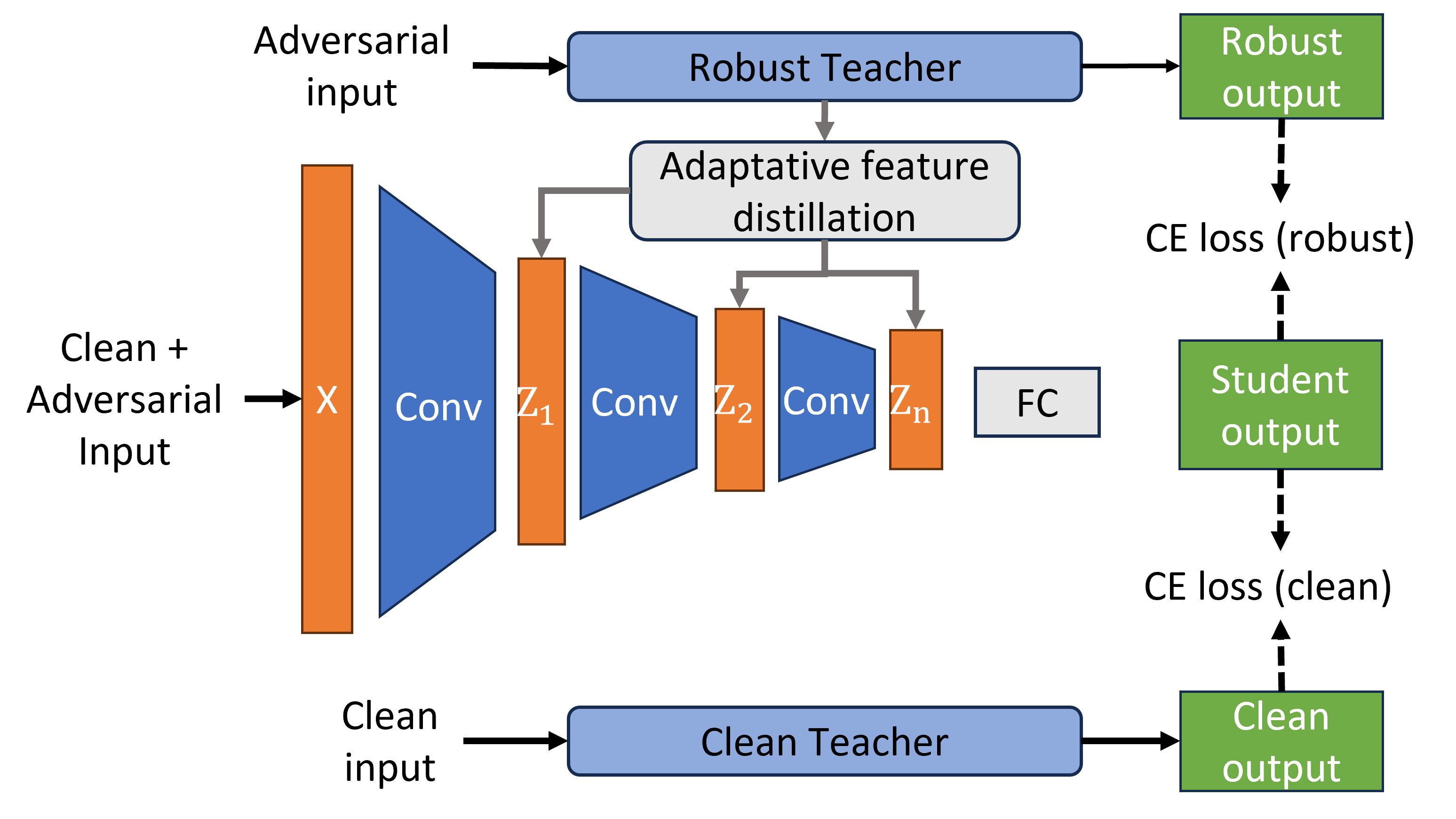}
    \caption{Diagram of the Double Distillation method. Clean samples are passed through the clean teacher and the student model, and the student model is trained by minimizing the cross entropy between the clean teacher outputs and its own. Adversarial samples are passed through the robust teacher and the student model, and the student model is trained by minimizing the cross entropy between these outputs and by minimizing the mean difference between attention weighted latent features.}
    \label{fig:twodist}
\end{figure}

\begin{figure}
    \centering
    \includegraphics[width=0.8\columnwidth]{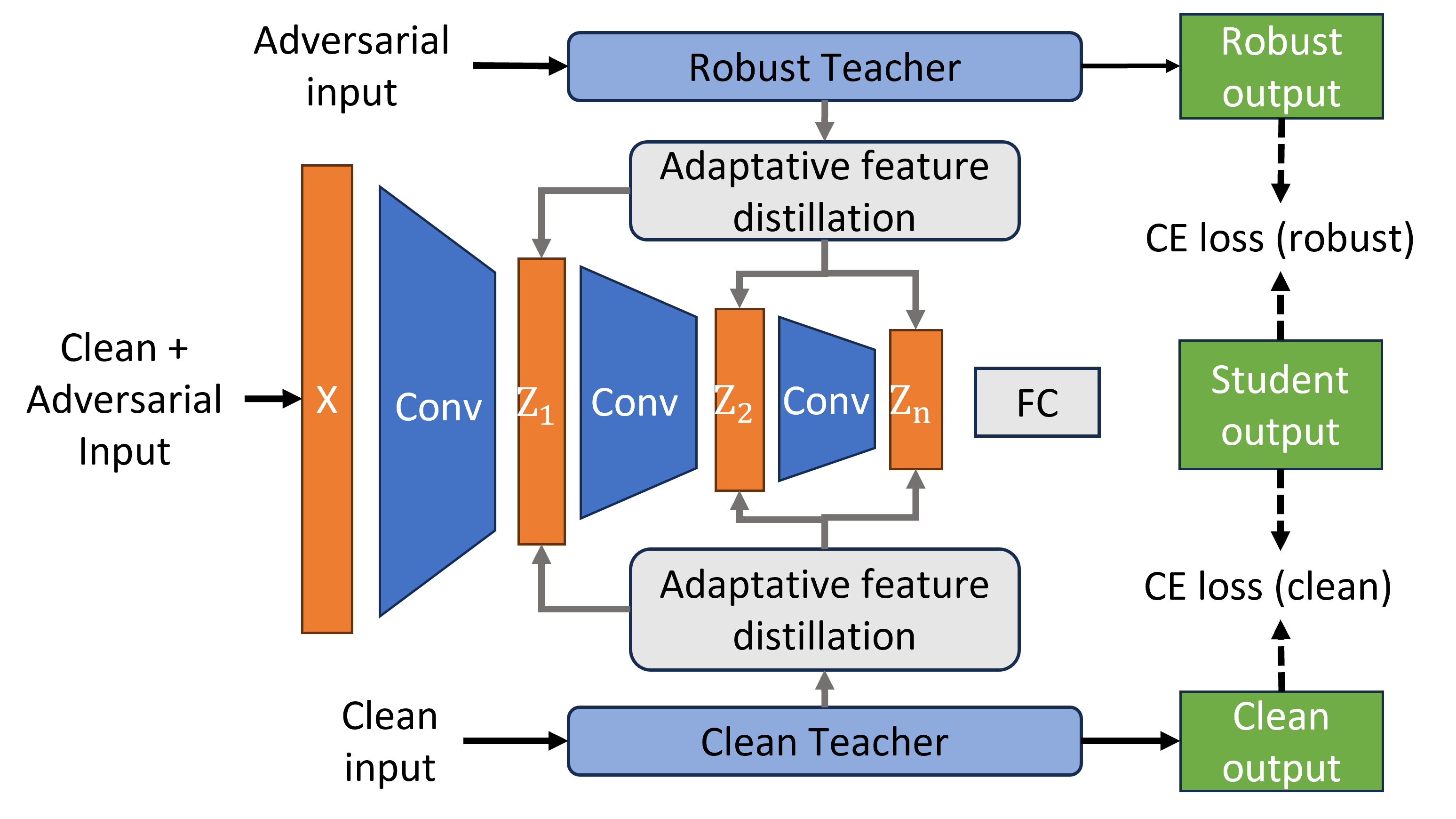}
    \caption{Diagram of the Joint Distillation Method. Clean samples are passed through the clean teacher and the student model. Adversarial samples are passed through the robust teacher and the student model. The student model is trained by minimizing the cross entropy between the teachers outputs and its own output, and by minimizing the mean difference between the teachers latent features and its own.}
    \label{fig:jointdist}
\end{figure}

The first method, Double Distillation (DD), illustrated in Figure \ref{fig:twodist}, uses a clean teacher to get soft labels from clean input and a robust teacher to get them from adversarial samples. The features distillation is performed, similarly to IBD, by using the robust teacher features alone. The loss function for the Double Distillation method:

\begin{equation}
\begin{split}
\mathcal{L}_{\text{double}} = \mathbb{E}_{p(x_{\text{nat}})} \Big[ & 
(1 - \alpha)\, \mathcal{L}_{\text{CE}}(h_s(x_{\text{nat}}), y_{\text{clean}}) \\
& + \alpha \, \mathcal{L}_{\text{CE}}(h_s(x_{\text{adv}}), y_{\text{rob}}) \\
& + \beta \sum_{i=1}^{n} \sum_{j=1}^{m} \text{Attn}_{i,j} 
\big(T_t^i(z_\text{rob}^i(x_{\text{adv}})) \\
& - T_s^j(z_s^j(x_{\text{adv}}))\big)^2
\Big]
\end{split}
\label{eq:double}
\end{equation}

\noindent where $y_{\text{rob}}$ and $y_{\text{clean}}$ correspond to the soft labels given by the robust and clean teachers, respectively. The second method, Joint Distillation (JD), is illustrated in Figure~\ref{fig:jointdist}. It considers both the clean teacher's features and the robust teacher's ones in the distillation, {\em i.e.}, both clean and robust teacher features are distilled into the student. The loss function for the Joint Distillation Method is:

\begin{equation}
\begin{split}
\mathcal{L}_{\text{joint}} = \mathbb{E}_{p(x_{\text{nat}})} \Big[ &
(1 - \alpha) \, \mathcal{L}_{\text{CE}}(h_s(x_{\text{nat}}), y_{\text{clean}}) \\
& + \alpha \, \mathcal{L}_{\text{CE}}(h_s(x_{\text{adv}}), y_{\text{rob}}) \\
& + \beta \Big( 
(1 - \alpha) \sum_{i=1}^{n} \sum_{j=1}^{m} \text{Attn}_{i,j}^{\text{clean}} 
\big(T_t^i(z_{\text{clean}}^i(x_{\text{nat}})) \\
& - T_s^j(z_s^j(x_{\text{nat}}))\big)^2 \\
& \quad + \alpha \sum_{i=1}^{n} \sum_{j=1}^{m} \text{Attn}_{i,j}^{\text{rob}} 
\big(T_t^i(z_{\text{rob}}^i(x_{\text{adv}})) \\
&- T_s^j(z_s^j(x_{\text{adv}}))\big)^2
\Big)
\Big]
\end{split}
\label{eq:joint}
\end{equation}

\noindent where $z^i_{\text{rob}}$ and $z^i_{\text{clean}}$ correspond to the latent features of the robust and clean teachers, respectively. Note that the attention weights $\text{Attn}_{i,j}^{\text{rob}}$ and $\text{Attn}_{i,j}^{\text{clean}}$ are learned separately for the two distillation terms.

\section{Experiments}

We evaluate the two proposed architectures through empirical experiments.  As a baseline, we used the performance of a reproduction of the IBD method reported in \cite{Kuang2023} and compared the performance of DD and JD.

\subsection{Experimental settings}

We use CIFAR-10 and CIFAR-100 datasets as our reference learning sets \cite{CIFAR10}. The student model was implemented using a Pre-Activation ResNet-18 network \cite{ResNet18}. For CIFAR-10, Both clean and robust teacher models are WideResNet-34x10 \cite{Zagoruyko2016}, and the robust teacher is trained using the TRADES method \cite{Zhang2019}. For the CIFAR-100 dataset, we use a WideResNet-22x6 \cite{Zagoruyko2016} for the clean teacher, and a WideResnet-34x10 \cite{Zagoruyko2016} trained using AWP \cite{Wu2020} as a robust teacher. Unless otherwise specified, the parameters are set to  $\alpha = 0.9$ and $\beta = 0.8$, matching the values used in IBD. All experiments are conducted on an NVIDIA H100 GPU. Adversarial samples are generated as in \cite{Kuang2023} using FGSM \cite{Goodfellow2015}, PGD \cite{Madry2018}, CW \cite{Carlini2017}, and Auto-Attack \cite{Croce2020}. The variation of parameter $\alpha$ is discussed in Appendix~\ref{sec:alpha} 

\subsection{Evaluation of the Accuracy/Robustness tradeoff}

Following \cite{Gowda2024}, we evaluate the model's overall performance using the harmonic mean to rigorously quantify the tradeoff between natural (clean) accuracy and adversarial robustness. The tradeoff score is calculated as follows:

\begin{equation}
     H(Acc_{\textrm{nat}}, Acc_{\textrm{rob}})= 2 \frac{Acc_{\textrm{nat}}Acc_{\textrm{rob}}}{Acc_{\textrm{nat}}+Acc_{\textrm{rob}}}
\end{equation}

where $\text{Acc}_{nat}$ denotes the accuracy on clean samples and $\text{Acc}_{rob}$ represents the robust accuracy under a given adversarial attack. We favor the harmonic mean over the arithmetic mean, as it effectively penalizes models that sacrifice robustness to artificially inflate clean accuracy.

\subsection{Comparison of accuracy}
\label{subsec:implementation}

We use the PGD attack with the $L^\infty$ radius of 8/255 and 10 steps with a step size of 2/255 to generate adversarial samples for training. All models are trained with the SGD optimizer, with a momentum of 0.9 and a batch size of 128. To achieve the best possible trade-off between standard accuracy and adversarial robustness, we test the proposed architectures across two different experimental configurations. The first setup, called ``global regularization",  uses an $L_2$ regularization valued at $5 \times 10^{-4}$ on all trained parameters and biases. For the second setup, called  ``selective regularization", we do not apply $L_2$ regularization to the bias terms in the student network nor to the layers in the Distillation module. The selective $L_2$ regularization alone leads to model overfitting and worse accuracy. In this setup, we increased the number of PGD steps used for adversarial training from 10 to 20. \newline

\noindent\textbf{Global regularization setup.} We have observed that in this setup, due to the high regularization, all the attention weights converge to the value of the reciprocal of the number of student features. The accuracies in this setup are shown in Table~\ref{tab:accuracy_global}. As the attention module becomes ineffective after the first dozen epochs (the weights become all equal), one may wonder whether attention is truly aiding in the training.  To test the usefulness of the attention module, we tried to suppress the attention module. The results of this experiment are shown in Table~\ref{tab:accuracy_attention}.

\begin{table}[htbp]
    \centering
    \caption{Robustness comparison with global regularization and pgd10 for training between the original Information Bottleneck Distillation Method and the proposed methods. The best results are \textbf{boldfaced}.}
    \label{tab:accuracy_global}
    \vspace*{3mm}
    \begin{tabular}{l c cccc} 
        \toprule
        \multirow{2}{*}{\textbf{Method}} & \multirow{2}{*}{\textbf{Clean}} & \multicolumn{4}{c}{\textbf{Attacks}} \\
        \cmidrule(lr){3-6}
        & & FGSM & PGD & CW & AA \\ 
        \midrule
        IBD  & 83.25 & 60.10 & 54.53 & 53.33 & 51.81 \\ 
        Double Distillation & \textbf{84.08} & 60.84  & 54.71 & 53.11& 51.69 \\ 
        Joint Distillations & 84.02 & \textbf{60.86} & \textbf{54.77} & \textbf{53.49} & \textbf{51.89} \\ 
        \bottomrule
    \end{tabular}
\end{table}

\begin{table*}[htbp]
    \centering
    \caption{Robustness comparison and Accuracy/Robustness tradeoff (Harmonic Mean) with and without attention module. The best results are \textbf{boldfaced}.}
    \label{tab:accuracy_attention}
    \vspace*{3mm}
    \resizebox{\textwidth}{!}{
    \begin{tabular}{ll c cccc cccc} 
        \toprule
        \multirow{2}{*}{\textbf{Attention}} & \multirow{2}{*}{\textbf{Method}} & \multirow{2}{*}{\textbf{Clean}} & \multicolumn{4}{c}{\textbf{Normal Values}} & \multicolumn{4}{c}{\textbf{Harmonic Mean}} \\
        \cmidrule(lr){4-7} \cmidrule(lr){8-11}
        & & & FGSM & PGD & CW & AA & FGSM & PGD & CW & AA \\ 
        \midrule
        \multirow{3}{*}{Yes} 
        & IBD & 83.25 & 60.10 & 54.53 & 53.33 & 51.81 & 69.81 & 65.90  & 65.01 & 63.87 \\ 
        & DD  & 84.08  & 60.84 & 54.71 & 53.11 & 51.69  & 70.60 & 66.29 & 65.10 & 64.02 \\ 
        & JD  & 84.02 & \textbf{60.86} & \textbf{54.77} & \textbf{53.49} & 51.89  & 70.59 & \textbf{66.31} & \textbf{65.37} & \textbf{64.16} \\ 
        \midrule
        \multirow{3}{*}{No} 
        & IBD & 83.25  & 60.41 & 54.63 & 53.41&\textbf{51.93} & 70.01 & 65.97 & 65.07 & 63.96 \\ 
        & DD  & \textbf{84.38} & 60.73 & 54.22 & 52.86& 51.22 & \textbf{70.62} & 66.02 & 65.00 & 63.75 \\ 
        & JD  & 84.04 & 60.19 & 54.22  & 52.88 & 51.36 & 70.14 & 65.91& 64.91  & 63.68 \\ 
        \bottomrule
    \end{tabular}
    }
\end{table*}

In the global regularization setup, where the values of the attention weights converge to the reciprocal of the number of student features, removing the attention module from the different methods causes the performances of DD and JD to drop for both clean and adversarial samples. The attention is useful for these methods, even if it is active only during the first epochs. Meanwhile, if we remove the attention module in this setup for the IBD method, there is no impact on clean accuracy, but robust accuracy increases. \newline

\begin{figure}
    \centering
    \includegraphics[width=0.45\columnwidth]{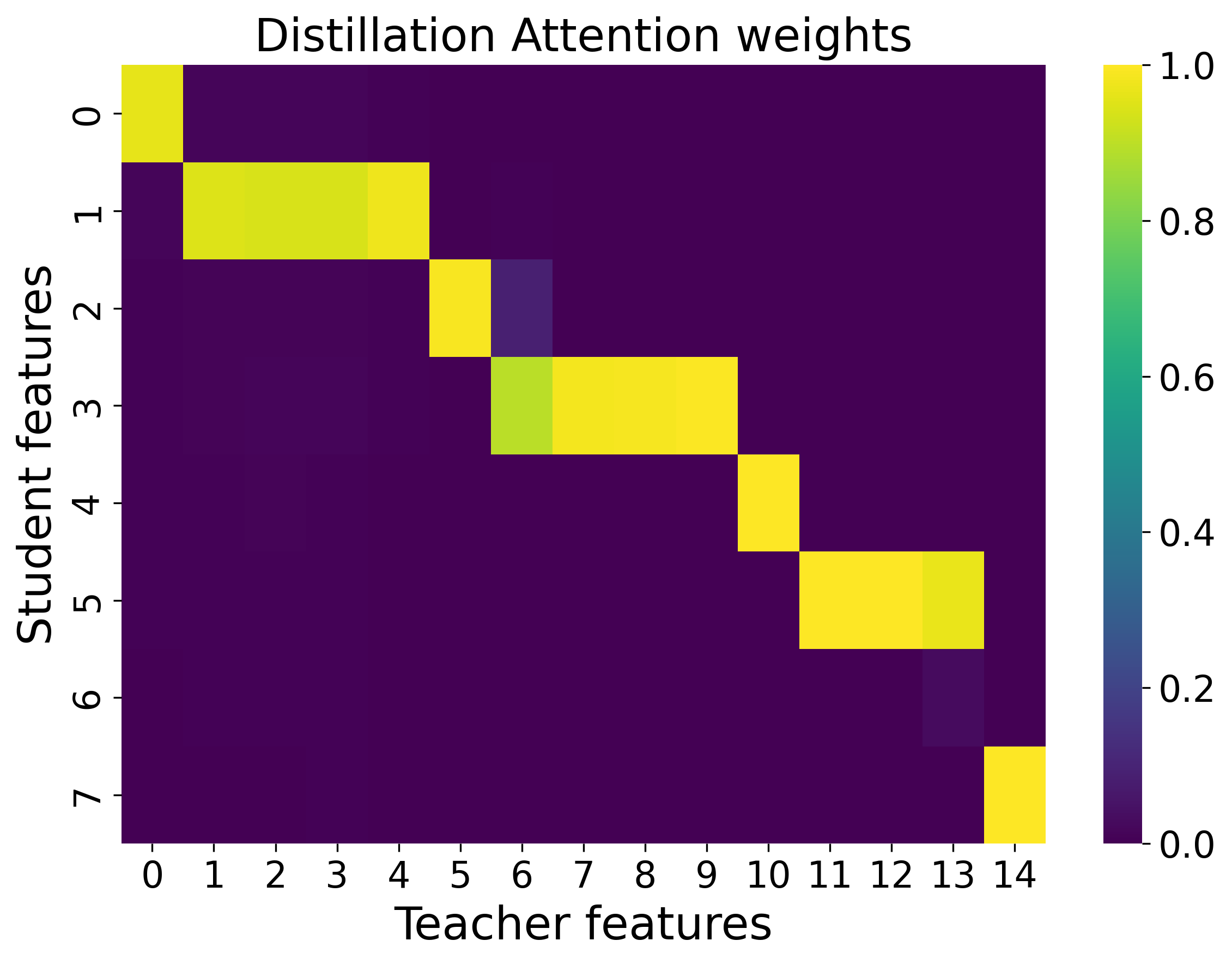}
    \caption{Attention weights in the Information Bottleneck Distillation}
    \label{fig:ibd_weights}
\end{figure}

\begin{figure}
    \centering
    \includegraphics[width=0.9\columnwidth]{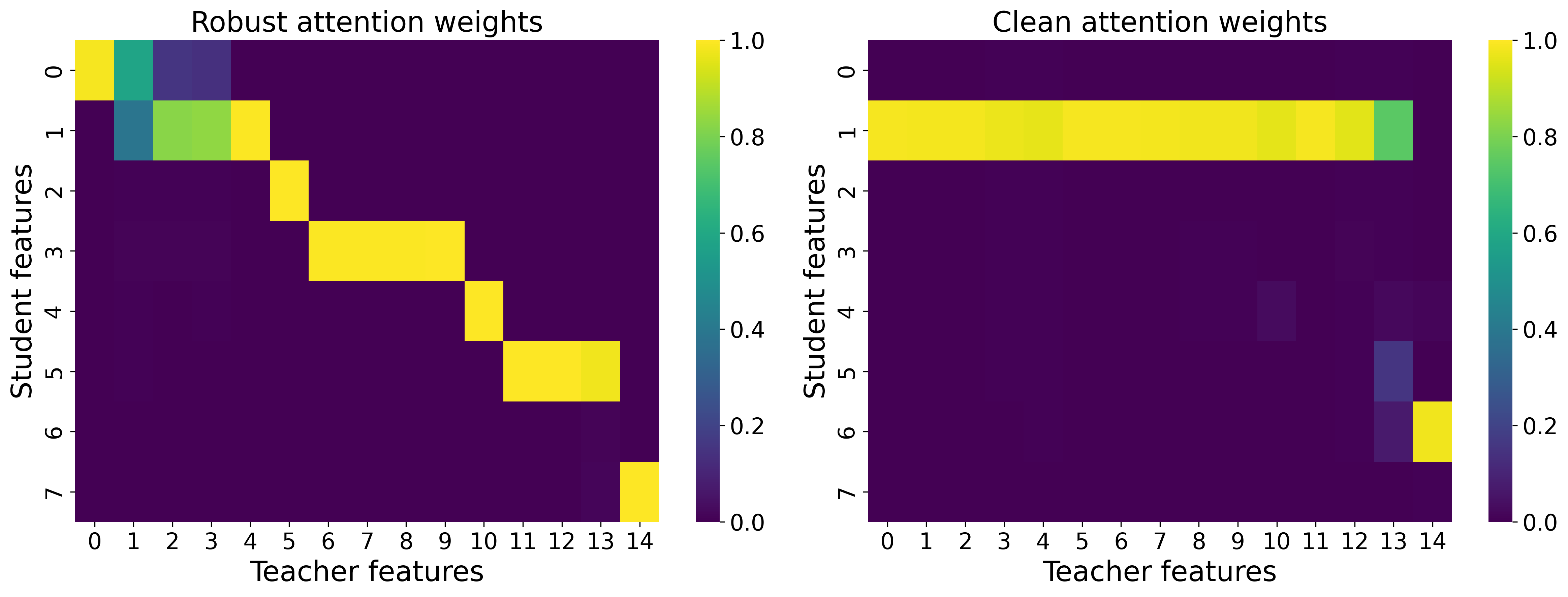}
    \caption{Attention weights in the Joint Distillation}
    \label{fig:jd_weigths}
\end{figure}

\noindent\textbf{Selective regularization setup.} Attention weights of IBD are shown in Figure~\ref{fig:ibd_weights}, and the attention weights for the distillation of the Joint Distillation are shown in Figure~\ref{fig:jd_weigths}. The accuracies for the three methods in this setup are shown in Table ~\ref{tab:accuracy_selective}. We can see that the distillation's attention weights between the robust teacher model and the student in the Joint Distillation model have a similar behavior to the attention weights of IBD. If we take a look at the clean distillation in the Joint Distillation model, we can see that the attention weights are focused on the second layer of the student model. \newline

\begin{table}[htbp]
    \centering
    \caption{Robustness comparison with selective regularization and pgd20 for training between the original Information Bottleneck Method and the proposed methods. The best results are \textbf{boldfaced}.}
    \label{tab:accuracy_selective}
    \vspace*{3mm}
    \begin{tabular}{l c cccc} 
        \toprule
        \multirow{2}{*}{\textbf{Method}} & \multirow{2}{*}{\textbf{Clean}} & \multicolumn{4}{c}{\textbf{Attacks}} \\
        \cmidrule(lr){3-6}
        & & FGSM & PGD & CW & AA \\ 
        \midrule
        IBD & 83.62  & 60.65 & \textbf{54.69} & \textbf{53.36} & \textbf{51.78} \\ 
        Double Distillation & 84.37 & \textbf{60.84} & 54.26& 52.93  & 51.40  \\ 
        Joint Distillations & \textbf{84.38} & 60.64 & 54.28 & 53.06  & 51.39 \\ 
        \bottomrule
    \end{tabular}
\end{table}

\noindent\textbf{Setups comparison.}  The comparison of the results in the two setups using the harmonic mean is shown in Table ~\ref{tab:accuracy_hmean}. From these results, we observe that the accuracy-robustness trade-off of IBD is more favorable in the selective regularization setup. Conversely, while DD and JD do not reach the performance of IBD under selective regularization, they demonstrate their maximum efficacy in the global regularization setup. Ultimately, when each method is evaluated in its respective optimal configuration, our proposed methods consistently achieve a higher harmonic mean than the baseline IBD.

\begin{table}[htbp]
    \centering
    \caption{Comparison by harmonic mean of the clean and robust accuracy between the original Information Bottleneck Distillation Method and the proposed methods. The best results are boldfaced.}
    \label{tab:accuracy_hmean}
    \vspace*{3mm}
    \begin{tabular}{ll c cccc} 
        \toprule
        \multirow{2}{*}{\textbf{Regularization}} & \multirow{2}{*}{\textbf{Method}} & \multirow{2}{*}{\textbf{Clean}} & \multicolumn{4}{c}{\textbf{Attacks}} \\
        \cmidrule(lr){4-7}
        & & & FGSM & PGD & CW & AA \\ 
        \midrule
        \multirow{3}{*}{Global} 
        & IBD & 83.25 & 69.81 & 65.90 & 65.01 & 63.87 \\ 
        & DD  & 84.08 & 70.60 & 66.29 & 65.10 & 64.02 \\ 
        & JD  & 84.02 & 70.59 & \textbf{66.31} & \textbf{65.37} & \textbf{64.16} \\ 
        \midrule
        \multirow{3}{*}{Selective} 
        & IBD & 83.62 & 70.30 & 66.13 & 65.15 & 63.96 \\ 
        & DD  & 84.06 & \textbf{70.63} & 66.08 & 65.20 & 64.09 \\ 
        & JD  & \textbf{84.38} & 70.57 & 66.06 & 65.15 & 63.88 \\ 
        \bottomrule
    \end{tabular}
\end{table}
 
\subsection{Impact of the size of the student model}

\begin{figure*}
    \centering
    \includegraphics[width=\textwidth]{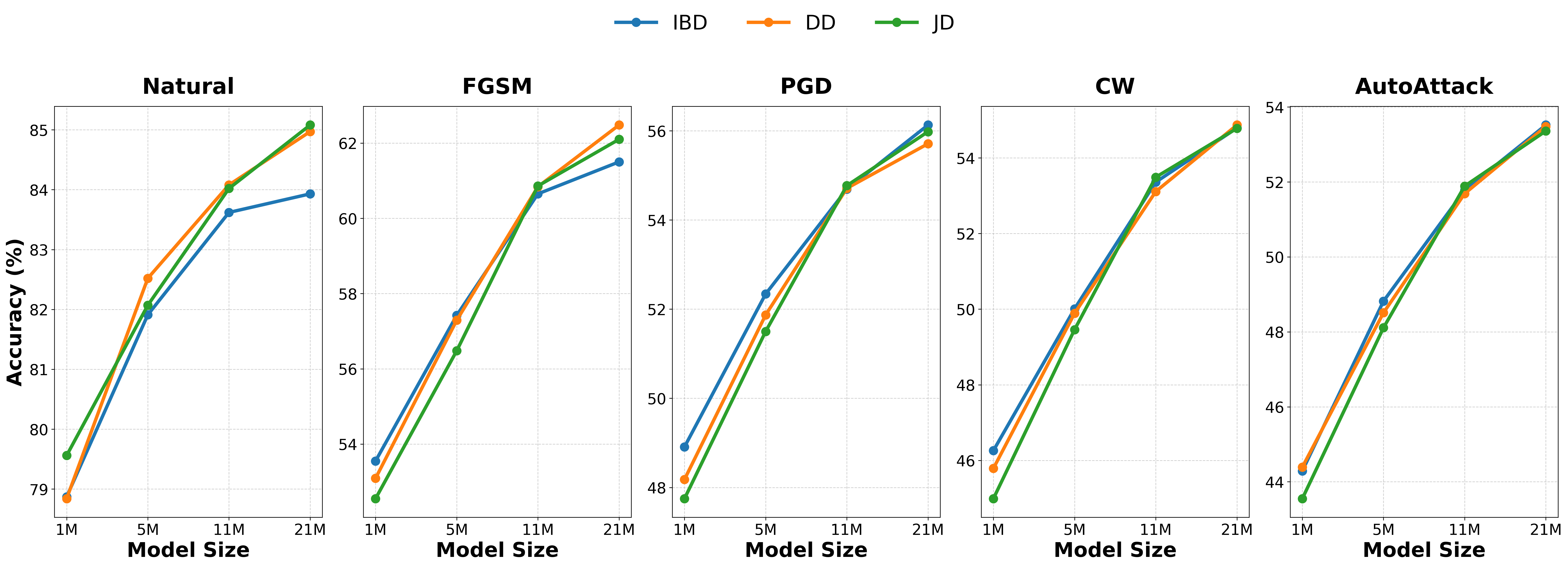}
    \caption{accuracy of the IBD and JD depending on the size of the student model (the y-axis scale is not the same in the different graphs)}
    \label{fig:accuracy_size}
\end{figure*}

In the basic studies, the teacher model is a WideResNet34, with 46 million parameters. In this section, we vary the size of the student model and see how it influences the model's accuracy.  We test the compression ratios between the student and the teacher to be 1/2, 1/4, 1/9, and 1/46. We use the best setup for each method. The results of this experiment are shown in Table~\ref{tab:accuracy_size_mean} and in Figure~\ref{fig:accuracy_size}. From these results, we can deduce that IBD performs best for small models, as robust accuracies of DD and JD drop faster than that of IBD. But for models with a size superior or equal to 1/4 of the size of the teacher model, the robust accuracy of DD and JD are very close and sometimes superior to the robust accuracy of the IBD method. Our methods, in which the student model receives information from two very different teachers, are more efficient as the student size increases and the student becomes more capable of dealing with a higher quantity of information. 

\begin{table*}[htbp]
    \centering
    \caption{Robustness comparison with student models of different sizes}
    \label{tab:accuracy_size_mean}
    \vspace*{3mm}
    \begin{tabular}{l c c cccc} 
        \toprule
        \multirow{2}{*}{\textbf{Method}} & \multirow{2}{*}{\textbf{Parameters}} & \multirow{2}{*}{\textbf{Clean}} & \multicolumn{4}{c}{\textbf{Attacks}} \\
        \cmidrule(lr){4-7}
        & & & FGSM & PGD & CW & AA \\ 
        \midrule
        \multirow{4}{*}{IBD} 
        & 1M  & 78.87 & 53.55 & 48.91 & 46.26 & 44.29 \\ 
        & 5M  & 81.91 & 57.42 & 52.34 & 50.01 & 48.82 \\ 
        & 11M & 83.62 & 60.65 & 54.69 & 53.36 & 51.78 \\ 
        & 21M & 83.93 & 61.50 & 56.13 & 54.79 & 53.52 \\ 
        \midrule
        \multirow{4}{*}{DD} 
        & 1M  & 78.84 & 53.09 & 48.18 & 45.79 & 44.39 \\ 
        & 5M  & 82.52 & 57.29 & 51.87 & 49.89 & 48.51 \\ 
        & 11M & 84.08 & 60.84 & 54.71 & 53.11 & 51.69 \\ 
        & 21M & 84.97 & 62.48 & 55.71 & 54.87 & 53.48 \\ 
        \midrule
        \multirow{4}{*}{JD} 
        & 1M  & 79.56 & 52.55 & 47.75 & 44.99 & 43.55 \\ 
        & 5M  & 82.07 & 56.48 & 51.50 & 49.46 & 48.11 \\ 
        & 11M & 84.02 & 60.86 & 54.77 & 53.49 & 51.89 \\ 
        & 21M & 85.08 & 62.10 & 55.98 & 54.78 & 53.36 \\ 
        \bottomrule
    \end{tabular}
\end{table*}

\subsection{Comparison with state-of-the-art methods}

In this section, we compare our methods against classical adversarial training (AT \cite{Madry2018}, TRADES \cite{Zhang2019}), robust distillation techniques (ARD \cite{Goldblum2020}, RSLAD \cite{Zi2021}), information bottleneck techniques (InfoAT \cite{Xu2022}, HBaR \cite{Wang2021}, IBD \cite{Kuang2023}), and a dual-teacher distillation technique (B-MTARD \cite{Zhao2024}).

Results are detailed in Table~\ref{tab:combined_robustness_cifar_both}. Overall, the proposed DD and JD architectures demonstrate highly competitive performance. On CIFAR-10, JD achieves the highest robust accuracy across all evaluated attacks. 
In addition, JD also achieves the highest harmonic mean under PGD, CW, and AA attacks, indicating a favorable balance between clean and robust accuracy. On CIFAR-100, while IBD yields slightly higher raw robustness under specific attacks (PGD, CW, AA), JD consistently secures the highest harmonic mean, indicating a more optimal trade-off between clean and robust accuracy. 

Appendix~\ref{sec:bmtard} discusses an extension of our method inspired by B-MTARD, where the balance between learning from the clean and robust teachers is adaptively adjusted during training instead of fixing the trade-off parameter \(\alpha=0.9\).

\begin{table*}[htbp]
    \centering
    \caption{Robustness comparison and Accuracy/Robustness tradeoff (Harmonic Mean) on CIFAR-10 and CIFAR-100}
    \label{tab:combined_robustness_cifar_both}
    \vspace*{3mm}
    \resizebox{\textwidth}{!}{
    \begin{tabular}{ll c cccc cccc}
        \toprule
        \multirow{2}{*}{\textbf{Dataset}} & \multirow{2}{*}{\textbf{Method}} & \multirow{2}{*}{\textbf{Clean}} & \multicolumn{4}{c}{\textbf{Accuracy}} & \multicolumn{4}{c}{\textbf{Harmonic Mean}} \\
        \cmidrule(lr){4-7} \cmidrule(lr){8-11}
        & & & FGSM & PGD & CW & AA & FGSM & PGD & CW & AA \\
        \midrule
        \midrule
        
        \multirow{12}{*}{\textbf{CIFAR-10}}
        & AT      & 83.40 & 58.14 & 51.47 & 50.39 & 47.76 & 68.52 & 63.66 & 62.82 & 60.74 \\
        & TRADES  & 83.51 & 59.42 & 53.33 & 50.91 & 49.75 & 69.43 & 65.09 & 63.26 & 62.35 \\
        \cmidrule(lr){2-11}
        & ARD     & 84.02 & 60.21 & 52.90 & 52.17 & 50.28 & 70.15 & 64.92 & 64.37 & 62.91 \\
        & RSLAD   & 83.40 & 59.99 & 54.69 & 52.89 & 51.49 & 69.78 & 66.06 & 64.73 & 62.91 \\
        \cmidrule(lr){2-11}
        & HBaR    & 85.07 & 58.60 & 52.80 & 50.63 & 49.75 & 69.49 & 65.25 & 63.56 & 62.86 \\
        & InfoAT  & 83.99 & 59.85 & 53.17 & 50.92 & 49.78 & 69.89 & 65.12 & 63.40 & 62.51 \\
        \cmidrule(lr){2-11}
        & B-MTARD & \textbf{89.52} & 60.31 & 50.91 & 49.27 & 45.12 & \textbf{72.07} & 64.91 & 63.56 & 60.00 \\
        \cmidrule(lr){2-11}
        & IBD     & 83.62 & 60.65 & 54.69 & 53.36 & 51.78 & 70.31 & 66.13 & 65.15 & 63.96 \\
        \cmidrule(lr){2-11}
        & DD      & 84.08 & 60.84 & 54.71 & 53.11 & 51.69 & 70.60 & 66.29 & 65.10 & 64.02 \\
        & JD      & 84.02 & \textbf{60.86} & \textbf{54.77} & \textbf{53.49} & \textbf{51.89} & 70.59 & \textbf{66.31} & \textbf{65.37} & \textbf{64.16} \\
        
        \midrule 
        \midrule
        
        \multirow{12}{*}{\textbf{CIFAR-100}}
        & AT & 56.21 & 33.46 & 31.34 & 28.03 & 25.48 & 41.95 & 40.24 & 37.41 & 35.07 \\
        & TRADES  & 57.37 & 32.44 & 30.37 & 26.50 & 24.94 & 41.44 & 39.72 & 36.25 & 34.77 \\
        \cmidrule(lr){2-11}
        & ARD     & 59.08 & 35.51 & 32.82 & 30.35 & 27.26 & 44.36 & 42.20 & 40.10 & 37.31 \\
        & RSLAD   & 58.05 & 35.59 & 33.62 & 30.38 & 28.39 & 44.13 & 42.58 & 39.89 & 38.13 \\
        \cmidrule(lr){2-11}
        & HBaR    & 59.09 & 33.38 & 30.56 & 26.85 & 25.25 & 42.66 & 40.29 & 36.92 & 35.38 \\
        & InfoAT  & 59.13 & 33.52 & 30.89 & 27.24 & 25.52 & 42.79 & 40.58 & 37.30 & 35.65 \\
        \cmidrule(lr){2-11}
        & B-MTARD & \textbf{65.25} & 34.10 & 28.94 & 26.00 & 23.06 & 44.83 & 40.13 & 37.21 & 34.10 \\
        \cmidrule(lr){2-11}
        & IBD     & 58.28 & 36.17 & \textbf{34.14} & \textbf{31.44} & \textbf{29.04} & 44.64 & 43.06 & 40.85 & 38.76 \\
        \cmidrule(lr){2-11}
        & DD      & 59.09 & 36.38 & 33.58 & 30.57 & 28.26 & 45.03 & 42.82 & 40.29 & 38.23 \\
        & JD      & 59.45 & \textbf{36.59} & 33.83 & 31.28 & 28.79 & \textbf{45.30} & \textbf{43.12} & \textbf{40.99} & \textbf{38.79} \\
        \midrule
        \bottomrule
    \end{tabular}
    }
\end{table*}

\subsection{Complexity of the different methods}

We have seen that the integration of the clean teacher improves clean accuracy without a significant drop in robust accuracy. The downside of these methods is the increase in training time and complexity. The original IBD method uses 260 minutes for training and 68.2M parameters. For DD and JD, as we add a teacher and a distillation module, we have around 115.4M parameters to train in about 1.5 times the IBD training time. Consequently, the choice of the method will be determined by the performance and time constraint trade-off. 

\section{Conclusion}

In this paper, we introduced two dual-teacher distillation methods based on the IBD framework. The Double Distillation method adds clean soft labels generated by the clean teacher, and the Joint Distillation method introduces soft labels for clean data and attention-weighted clean teacher's features distillation to the IBD method. We have evaluated these new architectures against the IBD baseline under two different regularization setups to determine the optimal accuracy-robustness trade-off for each method. The results demonstrate that our two proposed methods, when evaluated in their optimal configurations, consistently achieve a higher harmonic mean than the baseline IBD, indicating a better balance between clean and robust accuracy. Furthermore, we observed that these novel approaches are particularly effective for larger student models, whereas IBD maintains an advantage when applied to smaller capacity students.

\section*{Acknowledgments}
This work was supported by JSPS KAKENHI, Grant Number
JP23H05492, JP25K24611.


\appendix

\section{Technical appendices and supplementary material}

\subsection{Impact of $\alpha$}
\label{sec:alpha}

In all three methods, $\alpha$ represents the relative importance of clean training in comparison with adversarial training.  We show in Table ~\ref{tab:IBD_alpha} the impact of varying $\alpha$ from 0.85 to 0.95 for the three alternative methods: IBD, DD, and JD. We used the selective regularization setup for IBD and the global regularization setup for DD and JD.

\begin{table*}[htbp]
    \centering
    \caption{Impact of varying $\alpha$ in the alternative methods: IBD, DD, and JD. The best and second best values in each column are represented in \textbf{bold} and \textit{italic} style.}
    \label{tab:IBD_alpha}
    \vspace*{3mm}
    \resizebox{\textwidth}{!}{
    \begin{tabular}{l c c cccc cccc} 
        \toprule
        \multirow{2}{*}{\textbf{Method}} & \multirow{2}{*}{$\boldsymbol{\alpha}$} & \multirow{2}{*}{\textbf{Clean}} & \multicolumn{4}{c}{\textbf{Accuracy}} & \multicolumn{4}{c}{\textbf{Harmonic Mean}} \\
        \cmidrule(lr){4-7} \cmidrule(lr){8-11}
        & & & FGSM & PGD & CW & AA & FGSM & PGD & CW & AA \\ 
        \midrule
        \multirow{3}{*}{IBD} 
        & 0.85 & 83.65 & 60.52& 54.77 & 53.13 & \textbf{52.07} & 70.23 &66.20 & 64.99 & \textbf{64.19} \\ 
        & 0.9  & 83.62 & 60.65 & 54.69 & 53.36 & 51.78 & 70.31& 66.13 &65.15 & 63.96 \\ 
        & 0.95 & 83.19 & 60.12 & \textbf{54.78} & 53.17& 51.76& 69.80 & 66.06& 64.87& 63.81 \\ 
        \midrule
        \multirow{3}{*}{DD} 
        & 0.85 & \textit{84.32} & 60.49& 54.18& 52.51& 51.05 &70.44& 65.97 & 64.72& 63.60 \\ 
        & 0.9  & 84.08& \textit{60.84} & 54.71 & 53.11& 51.69 & \textbf{70.60} & \textit{66.29} & 65.10 & 64.02 \\ 
        & 0.95 & 83.69& 60.48& 54.30& 53.11& 51.45 & 70.22 & 65.86& 64.98 & 63.72 \\ 
        \midrule
        \multirow{3}{*}{JD} 
        & 0.85 & \textbf{84.34} & 60.65& 54.46& \textit{53.42} & 51.73 & 70.56 & 66.17  & \textbf{65.40} & 64.13 \\ 
        & 0.9  & 84.02 & \textbf{60.86} & \textit{54.77} & \textbf{53.49} & \textit{51.89} & \textit{70.59} & \textbf{66.31} & \textit{65.37} & \textit{64.16} \\ 
        & 0.95 & 83.48 & 60.36 & 54.45 & 53.21 & 51.63  & 70.06 & 65.91 & 64.99& 63.79 \\ 
        \bottomrule
    \end{tabular}
    }
\end{table*}

A closer inspection of the results reveals several key findings regarding the models' behavior. First, concerning standard performance, setting $\alpha=0.85$ consistently yields the highest clean accuracy across all three methods. More importantly, the proposed DD and JD architectures maintain a distinct advantage in standard accuracy over the IBD baseline across the entire $\alpha$ spectrum (e.g., reaching up to 84.34\% for JD compared to a maximum of 83.65\% for IBD). This validates our initial hypothesis: the introduction of a clean teacher effectively helps the student preserve standard features that are typically degraded during adversarial training. 

Second, regarding adversarial robustness, the sensitivity to $\alpha$ diverges between the baseline and our proposed methods. While IBD achieves its highest AutoAttack (AA) robustness at $\alpha=0.85$, its performance on other attacks remains inconsistent across the varying values. In contrast, both DD and JD demonstrate a clear peak in adversarial defense when $\alpha$ is set to 0.9. At this specific configuration, JD secures the highest or second-highest robust accuracy against FGSM, PGD, CW, and AA simultaneously, demonstrating a more stable defensive profile.

Finally, evaluating the overall trade-off through the harmonic mean confirms the efficacy of this configuration. Across this spectrum, the proposed JD method exhibits a highly predictable and smooth behavior. Rather than achieving singular extreme values on isolated metrics, JD with $\alpha=0.9$ offers a reliable "sweet spot". It yields a balanced harmonic mean that consistently ranks at the top across all attack evaluations, proving that the joint distillation process successfully harmonizes standard and robust accuracies without critical compromises.

\subsection{Implementing B-MTARD logic}
\label{sec:bmtard}

Presented in \cite{Zhao2024}, B-MTARD method consists of two distillations, from a robust teacher and a clean teacher to a student model. The distillation loss function is as follows:

\begin{equation}
    \mathcal{L}_\textrm{B-MTARD}=w_\textrm{nat}\mathcal{L}_\textrm{CE}(h(x_\textrm{nat},\tau_\textrm{s}), y_\textrm{clean}(x_\textrm{nat},\tau_\textrm{clean})) + w_\textrm{adv}\mathcal{L}_\textrm{CE}(h(x_\textrm{adv},\tau_\textrm{s}), y_\textrm{rob}(x_\textrm{adv},\tau_\textrm{rob}))
\end{equation}

with $\tau_\textrm{s}$ the temperature of the output softmax of the student model, and $\tau_\textrm{clean}$ and $\tau_\textrm{rob}$ the temperatures of the output softmax of the clean
and robust teacher, respectively. B-MTARD balances the learning speed of the student model by balancing the loss weights $w_\textrm{nat}$ and $w_\textrm{adv}$. 

\begin{equation}
    w_\textrm{nat}(t) = r_\textrm{w}r_\textrm{nat}(t) + (1-r_\textrm{w})w_\textrm{nat}(t-1)
\end{equation}
\begin{equation}
    w_\textrm{adv}(t) = r_\textrm{w}r_\textrm{adv}(t) + (1-r_\textrm{w})w_\textrm{adv}(t-1)
\end{equation}

with $r_\textrm{w} = 0.025$ the weight learning rate.  

B-MTARD also balances the amount of information received by both teachers by balancing the teachers' outputs' softmax temperature. If one teacher’s output probability distribution is sharper than the others, increasing the temperature will soften its predictions, producing a smoother probability distribution and reducing overconfidence, which facilitates a more balanced knowledge transfer during distillation. The temperatures of both teacher's output softmax are updated as follows:

\begin{equation}
    \tau_\textrm{clean} = \tau_\textrm{clean} - r_\tau \, \textrm{sign}\Big(H(y_\textrm{clean}) - H(h_\textrm{s}(x_\textrm{nat}))\Big)
\end{equation}
\begin{equation}
    \tau_\textrm{rob} = \tau_\textrm{rob} - r_\tau \, \textrm{sign}\Big(H(y_\textrm{rob}) - H(h_\textrm{s}(x_\textrm{adv}))\Big)
\end{equation}

with $H$ being the information entropy and $r_\tau =  0.001$ the temperature update rate.

We tried to implement the B-MTARD loss weight and temperature balance in both of our proposed methods, DD and JD, and called them Balanced Double Distillation (B-DD) and Balanced Joint Distillation (B-JD). The results are shown in Table~\ref{tab:accuracy_balanced}.

\begin{table*}[htbp]
    \centering
    \caption{Robustness comparison of our balanced methods.}
    \label{tab:accuracy_balanced}
    \vspace*{3mm}
    \resizebox{\textwidth}{!}{
    \begin{tabular}{l c cccc cccc} 
        \toprule
        \multirow{2}{*}{\textbf{Method}} & \multirow{2}{*}{\textbf{Clean}} & \multicolumn{4}{c}{\textbf{Accuracy}} & \multicolumn{4}{c}{\textbf{Harmonic Mean}} \\
        \cmidrule(lr){3-6} \cmidrule(lr){7-10}
        & & FGSM & PGD & CW & AA & FGSM & PGD & CW & AA \\ 
        \midrule
        DD   & 84.08 & 60.84 & 54.71 & 53.11 & 51.69 & 70.60 & 66.29 & 65.10 & 64.02 \\ 
        JD   & 84.02 & \textbf{60.86} & \textbf{54.77} & \textbf{53.49} & \textbf{51.89} & 70.59 & 66.31 & \textbf{65.37} & \textbf{64.16} \\ 
        \midrule
        B-DD & 85.12 & 60.32 & 54.4 & 52.89 & 50.37 & \textbf{70.61} & \textbf{66.38} & 65.24 & 63.29 \\
        B-JD & \textbf{86.77} & \textbf{60.86} & 52.55 & 51.15 & 47.14 & 71.54 & 65.46 & 64.36 & 61.09 \\
        \bottomrule
    \end{tabular}
    }
\end{table*}

It can be observed that incorporating these balancing mechanisms alters the learning dynamics of both methods, biasing them towards clean samples. We chose to retain the original DD and JD as our primary methods for this paper because the balanced variants exhibit a performance drop against strong adversarial attacks (notably AutoAttack), which negatively impacts the overall robustness-accuracy tradeoff.

\vspace{12pt}


\end{document}